\documentclass[journal]{IEEEtran}
\usepackage{url}
\usepackage{amsmath,amssymb}
\usepackage{algorithm}
\usepackage{algorithmic}
\usepackage{graphicx}
\usepackage{epstopdf}
\usepackage{amsmath}
\usepackage{amssymb}
\usepackage{multirow}

\usepackage{algorithm}
\usepackage{algorithmic}
\usepackage{amsfonts}

\begin{document}

\title{Implicitly Incorporating Morphological Information into Word Embedding}

\author{
\IEEEauthorblockN{Yang Xu$^*$ and Jiawei Liu$^*$}
\thanks{* The two authors contributed equally to this paper.}\\
\IEEEauthorblockA{School of Computer Science and Technology\\
University of Science and Technology of China\\
Hefei, China 230027\\
Email: \{smallant, ustcljw\}@mail.ustc.edu.cn}
}

\maketitle

\begin{abstract}
In this paper, we propose three novel models to enhance word embedding by implicitly using morphological information. Experiments on word similarity and syntactic analogy show that the implicit models are superior to traditional explicit ones. Our models outperform all state-of-the-art baselines and significantly improve the performance on both tasks. Moreover, our performance on the smallest corpus is similar to the performance of CBOW on the corpus which is five times the size of ours. Parameter analysis indicates that the implicit models can supplement semantic information during the word embedding training process. \end{abstract}


\section{Introduction}
Distributed word representation, which is also called word embedding, has been a hot topic in the area of Natural Language Processing (NLP). For its effectiveness, the learned word representations have been used in plenty of tasks such as text classification \cite{liu2015topical}, information retrieval \cite{manning2008introduction} and sentiment analysis \cite{Shin2016Lexicon}. There are many classic word embedding methods including the Semantic Extraction using a Neural Network Architecture (SENNA) \cite{Collobert2007Fast}, Continuous  Bag-of-Word (CBOW), Skip-gram \cite{mikolov2013efficient}, Global Vector (Glove) \cite{pennington2014glove}, etc. These models can only capture word-level semantic information and ignore the meaningful inner structures of words such as English morphemes and Chinese characters.

The effectiveness of exploiting the internal structures has been validated by several models \cite{Chen2015Joint,Xu2016Improve,luong2013better}. These models split a word into several parts, which are directly added into the input layer of their models to improve the representation of target word. In this paper, we call these methods which directly utilize the internal structures as \emph{explicit} models. Actually, incorporating the inner information of words into word embedding is meaningful and promising. For English morphemes, words starting with prefix ``a" or ``an" will have the meanings of ``not" and "without" like ``asexual" and ``anarchy". In addition, words ending with suffix ``able" or ``ible" will have the meaning of ``capable" like ``edible" and ``visible". Obviously, both the morpheme and their meanings are beneficial. Nevertheless, the explicit models only care about morphemes themselves and ignore their meanings. In vector space of explicit model,  morpheme-similar words will have a tendency to stay together while the long distances between these words and their morphemes' meanings are still remained. For clarity, we illustrate the basic idea of these models in the left part of Figure 1. 
\begin{figure}[!t]
	\centering
	\includegraphics[width=1.0\linewidth]{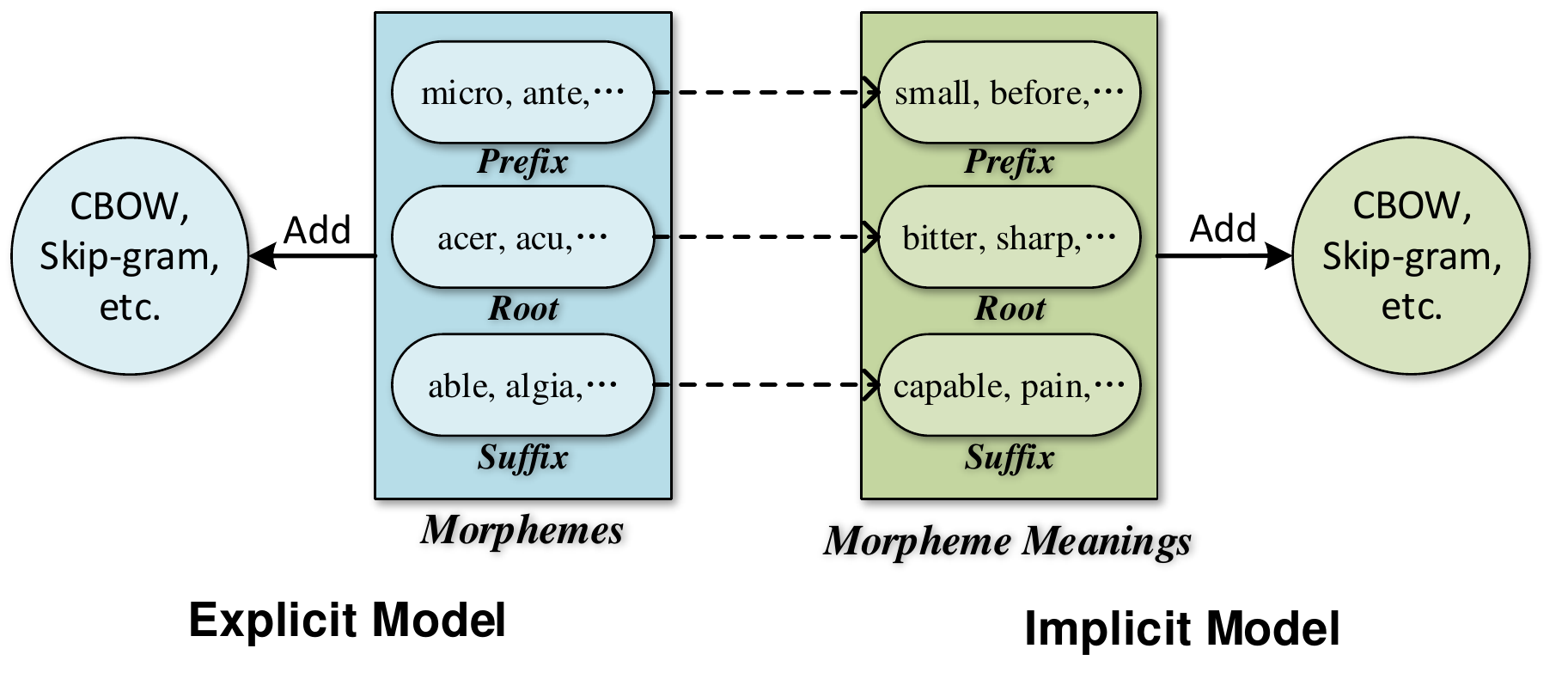}
	\caption{Basic ideas of traditional morpheme-enhanced models and our models. The left figure illustrates tradition models where the morpheme information like prefix ``micro" is directly add to some classic models. Our idea is shown in the right part where the corresponding positions of morphemes are replaced by their meanings.}
\end{figure}

In this paper, we propose three more refined word embedding models by implicitly exploiting the meanings of morphemes rather than morphemes themselves. Based on our knowledge, morphemes used in this paper are divided into 3 parts including prefix, root and suffix. In the vector space of our models, morpheme-similar words in vector space will not only have a tendency to gather together but also tend to locate close to their morphemes' meanings. The basic ideas of our methods are shown in the right part of Figure 1. For comparison, our models together with the state-of-the-art baselines are tested on word similarity and syntactic analogy. The result shows that our models outperform the baselines and get significantly improvement on both tasks. The contributions of this paper are summarized as follows.
\begin{itemize}
\item To the best of our knowledge, we are the first to implicitly utilize the morphological information. Based on this strategy, word embeddings in vector space have a more precise structure, which make our models significantly outperform the state-of-the-art baselines on experiments.
\item According to parameter analysis, our models seem to have the ability to  increase the semantic information in the corpus, which lead us to believe our models are of great advantages to deal with some morpheme-rich but low-resource languages.
\item The strategy of implicitly incorporating morphological information can benefit the areas where the morphemes are directly utilized in the past. For convenience to researchers interested in this area, we create a word map where the key is a word and the value is its morphemes' meanings set.
\end{itemize}


\section{Background and Related Work}
Our implicit models are built upon the CBOW \cite{mikolov2013efficient}. We firstly review the backgrounds of CBOW, and then present related work on recent word-level and inner structure-based word embedding methods.

\textbf{CBOW with Negative Sampling}\
With a slide window, CBOW \cite{mikolov2013efficient} utilizes the context words in the window to predict the target word. Given a sequence of tokens $T=\{t_1,t_2,\cdots,t_n\}$, the objective of CBOW is to maximize the following average log probability equation:
\begin{equation}
\label{cbow}
L=\frac{1}{n}\sum_{i=1}^n\log\ p(t_i|context(t_i)),
\end{equation}
where $context(t_i)$ means the context words of $t_i$ in the slide window. Based on softmax, $p(t_i|context(t_i))$ is defined as follows:
\begin{equation}
\label{cbow1}
\frac{\exp(vec^{'}(t_i)^{T}\sum_{-k\leq j\leq k,j\neq 0}vec(t_{i+j}))}{\sum^V_{x=1}\exp(vec^{'}(t_x)^T\sum_{-k\leq j\leq k,j\neq 0}vec(t_{i+j}))},
\end{equation}
where $vec(t)$ and $vec^{'}(t)$ are the ``input" and ``output" vector representation of token $t$, $k$ means the slide window size and $V$ means the vocabulary size. Due to huge size of English vocabulary, however, Eq.(\ref{cbow1}) can not be calculated in a tolerable time. Therefore, negative sampling and hierarchical softmax are proposed to solve this problem. Due to the efficiency of negative sampling, all related models are trained based on it. In terms of negative sampling, every item whose form is like $\log p(t_O|t_I)$ is defined as follows:
\begin{equation}
\label{negative}
\begin{split}
&\log\delta(vec^{'}(t_O)^Tvec(t_I))+\\
&\sum^m_{i=1}\log[1-\delta(vec^{'}(t_i)^Tvec(t_I))],
\end{split}
\end{equation}
where $m$ indicates the number of negative samples, and $\delta(\cdot)$ is the sigmoid function. The first item of Eq.(\ref{negative}) is the probability of target word when its context is given. The second item  Eq.(\ref{negative})  denotes the probability that negative samples don't share the same context as the target word.

\textbf{Word-Level Word Embedding}\
In general, word embedding models are divided into two main branches. One is based on neural network while the other is based on matrix factorization. Except for the CBOW, Skip-gram is another widely used model, which predicts the context by using the target word \cite{mikolov2013efficient}. Based on matrix factorization, Dhillon et al. proposed a spectral word embedding method to measure the correlation between word information matrix and context information matrix \cite{dhillon2015eigenwords}. In order to combine the advantage of neural network-based model with the advantage of matrix factorization, Pennington proposed another famous word embedding model named Glove, which is reported to outperform the CBOW and Skip-gram model \cite{mikolov2013efficient} on some tasks \cite{pennington2014glove}. These models are all effective to capture word-level semantic information while neglect inner information of words. On the contrast, the unheeded inner inner information is utilized in both our implicit models and other explicit models.

\textbf{Inner Structure-based Word Embedding}
Recently, some more fine-grained word embedding models are proposed by exploiting the internal structures. Chen et al. proposed a character-enhanced Chinese word embedding model, which split a Chinese word into several characters and add the characters into the input layer of their models \cite{Chen2015Joint}. Based on recursive neural network, Luong et al. utilized the weighted morpheme part and basic part of an English word to improve the representation of it \cite{luong2013better}. Kim et al utilized the convolutional character information to embed the English words\cite{Kim2015Character}. Their model can learn semantic information from character-level, which is proved to be effective to deal with some morpheme-rich languages. With a huge size architecture, however, it is very time-consuming. Cotterell et al. augmented the log linear model to make the words, which share morphological information, gather together in vector space \cite{Cotterell2015Morphological}. These models can improve the linguistic similarity with respect to semantic, syntactic or morphology to some degree. Nevertheless, these explicit models can only capture the facial linguistic information of internal structures and ignore their meanings. In contrast, our models not only exploit the inner information but also consider about the valuable meanings of inner structures.

\section{Implicit Morpheme-Enhanced Word Embedding}

In this section, we present 3 novel models named MWE-A, MWE-S and MWE-M by implicitly incorporating the morphological information. MWE-A assumes that all morphemes' meanings of a token have equal contributions to the representation of the target word. MWE-A is applicable to the refined morphemes' meanings set which contains  a few outliers. However, refining the morphemes' meanings set is time-consuming and needs a lot of human annotation, which costs a lot. MWE-S is proposed to solve this problem. Motivated by the attention scheme, MWE-S holds the assumption that all morphemes' meanings have different contributions. In this model, the outliers are weighed with a small value, which means they will have a few effects on the representation of the target word. Different from MWE-A and MWE-S,  MWE-M actually is an approximate method. In this model, we only care about the morphemes' meanings which have the greatest contribution to the target word. The remaining words are viewed as outliers and removed out of the morphemes' meanings set. We also discussed the strategy to find the words' morphemes in this section.

\begin{figure}[!t]
	\centering
	\includegraphics[width=1.0\linewidth]{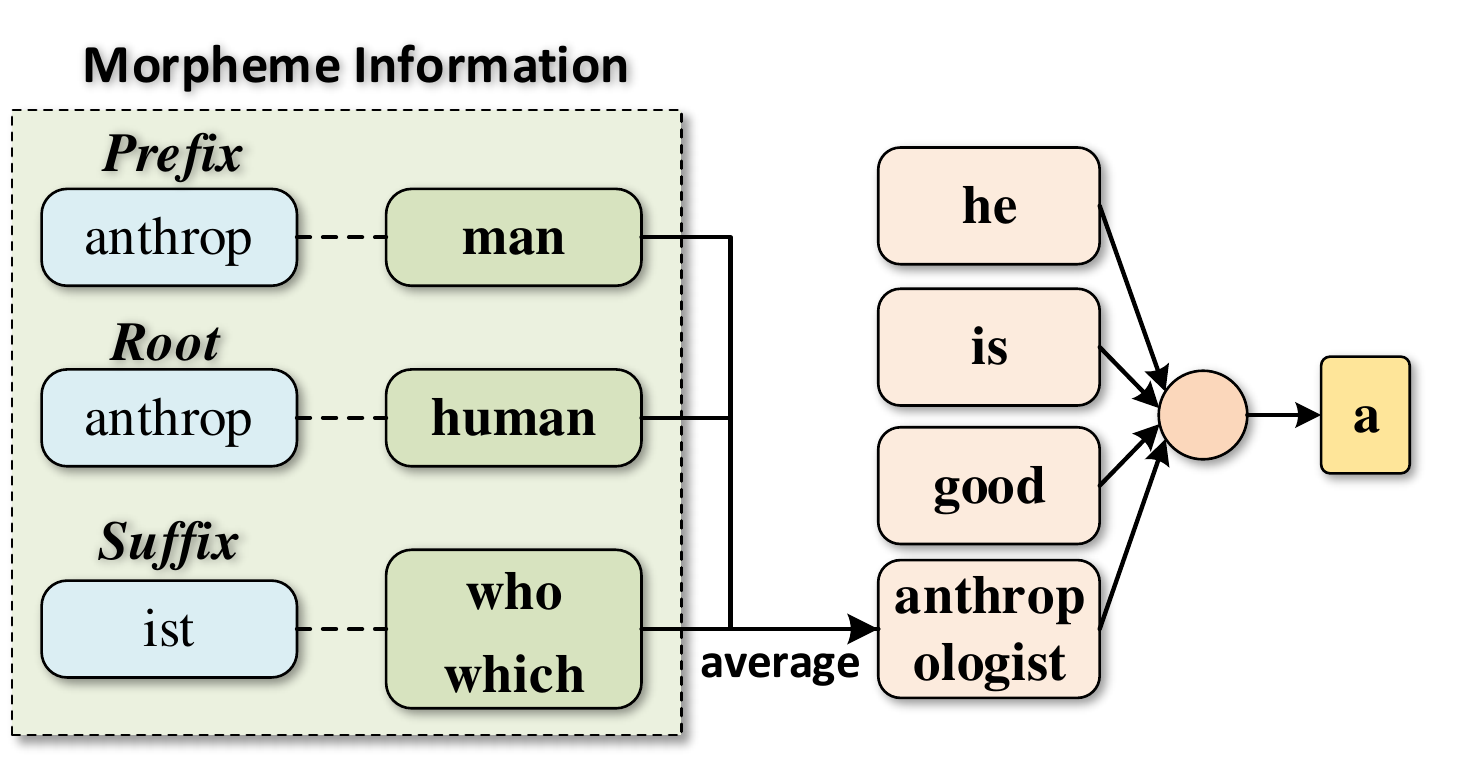}
	\caption{This is figurative description of MWE-A. For clarity, we utilize the sentence ``he is a good anthropologist" as an example. When calculating the input vector of ``anthropologist", we split the word into prefix, root and suffix. Prefix ``anthrop", root ``anthrop" and suffix ``ist" have the meaning of ``man", ``human" and ``who, which" respectively. The representations of all morphemes' meanings are added to the representation of ``anthropologist" with same weight.}
\end{figure}

\subsection{Morpheme-Enhanced Word Embedding-Average (MWE-A)}
This model is built on the assumption that all morphemes' meanings of token $t_i$ have equal contributions to $t_i$. Given a sequence of tokens $T=\{t_1,t_2,\cdots,t_n\}$, we assume that the morphemes' meanings set of $t_i (i\in[1,n])$ is $M_i$. $M_i$ can be divided into three parts $P_i$, $R_i$ and $S_i$, which indicate the prefix meaning set, root meaning set and suffix meaning set of $t_i$, respectively. Hence, when $t_i$ is the context word of $t_j$, the modified embedding of $t_i$ can be defined as follows:
\begin{equation}
\widehat{v}_{t_i}=\frac{1}{2}(v_{t_i}+\frac{1}{N_i}\sum_{w\in M_i}v_w),
\end{equation}
where $v_{t_i}$ is the original word embedding of $t_i$, $N_i$ denotes the length of $M_i$ and $v_w$ indicates the vector of $w$. In this paper, we assume the original word embedding and the sum of its morphemes' meanings embedding have equal weight. Namely, they are both 0.5. More details can be found in Figure 2.

\subsection{Morpheme-Enhanced Word Embedding-Similarity (MWE-S)}
This model is built based on the attention scheme. We observe that a lot of words have more than one morpheme's meanings. For instance, word ``anthropologist", its morpheme's meaning set is \{man,human,who,which\}. Obviously, ``man" and ``human" are closer to ``anthropologist". Motivated by this observation, we assume that different morphemes have different contributions. In this paper, we firstly train a part of word embeddings based on CBOW. Then, based on the pre-trained word embedding, the different contributions can be measured by calculating the cosine distances between target word and its morphemes' meanings. We illustrate the main idea of this model in Figure 3. The notation in MWE-A is reused to describe MWE-S. Mathematical formation is given by
\begin{equation}
\label{MWE-S}
\widehat{v}_{t_i}=\frac{1}{2}[v_{t_i}+\sum_{w\in M_i}v_w\ sim(v_{t_i},v_w)],
\end{equation}
where $v_{t_i}$ is the original vector of $t_i$, $sim(v_{t_i},v_w)$ is the function used to denote the similarities between $t_i$ and its morphemes' meanings. We define the function $cos(v_i,v_j)$ to denote the cosine distance between $v_i$ and $v_j$. Then, $sim(v_{t_i},v_w)$ is normalized as follows:
\begin{equation}
\label{normalize}
sim(v_{t_i},v_w)=\frac{cos(v_{t_i},v_w)}{\sum_{w\in M_i}cos(v_{t_i},v_w)},\ w\in M_i,
\end{equation}

\begin{figure}[!t]
	\centering
	\includegraphics[width=1.0\linewidth]{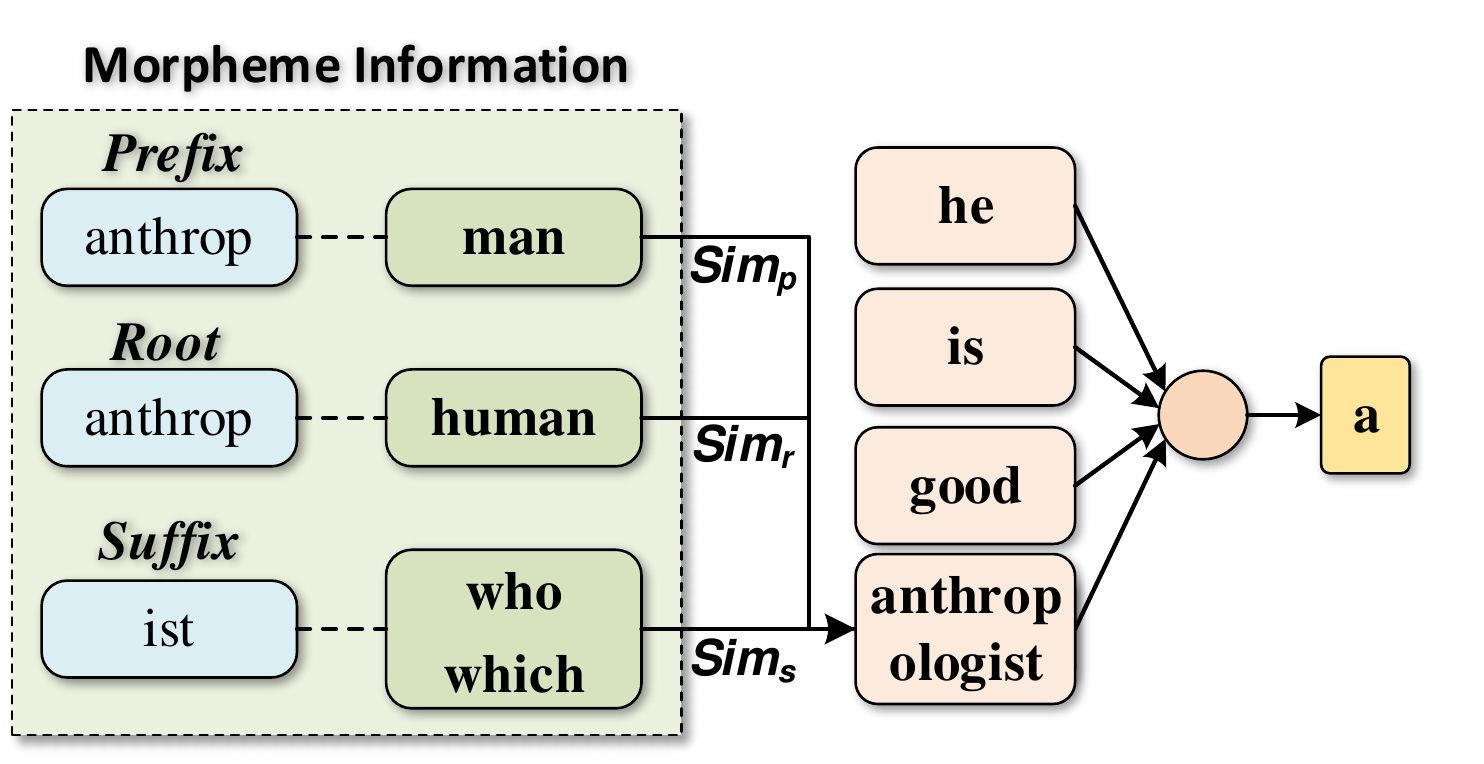}
	\caption{MWE-S. We also use the sentence mentioned above as an example. In this model, all morphemes' meanings are added with different weights.}
\end{figure}

\subsection{Morpheme-Enhanced Word Embedding-Max (MWE-M)}
From the collections of morphemes and their meanings, we find that except for multiple meanings of a morpheme, the corresponding meanings of a token's prefix, root and suffix are usually different.  However, some meanings are not close to the target word, which may affect the quality of word representations. To achieve better performance, we only exploit the morpheme's meaning whose distance to target word is the max. More details are shown in Figure 4. In that figure, the suffix's meanings of ``anthropologist" are ``who" and ``which". According to our description, the word ``which" in red will not be applied for its shorter distance to ``anthropologist". For a single token $t_i$, the new morphemes' meanings set is defined as $M_{max}^i=\{P_{max}^i,R_{max}^i,S_{max}^i\}$ where $P_{max}^i,R_{max}^i, S_{max}^i$ are defined as follows:
\begin{eqnarray}
  \nonumber P_{max}^i &=& arg\max\limits_{w}cos(v_{t_i},v_w), w\in P_i, \\
  R_{max}^i &=& arg\max\limits_{w}cos(v_{t_i},v_w), w\in R_i, \\
  \nonumber S_{max}^i &=& arg\max\limits_{w}cos(v_{t_i},v_w), w\in S_i,
\end{eqnarray}
where $cos(\cdot)$ stands for cosine distance. Hence, MWE-M is defined mathematically as follows:
\begin{equation}
\label{MWE-M}
\widehat{v}_{t_i}=\frac{1}{2}[v_{t_i}+\sum_{w\in M_{max}^i}v_w\ sim(v_{t_i},v_w)],
\end{equation}
where $sim(\cdot)$ indicates the similarity between two words and needs to be normalized like Eq.(\ref{normalize}) shows.

\begin{figure}[!t]
	\centering
	\includegraphics[width=1.0\linewidth]{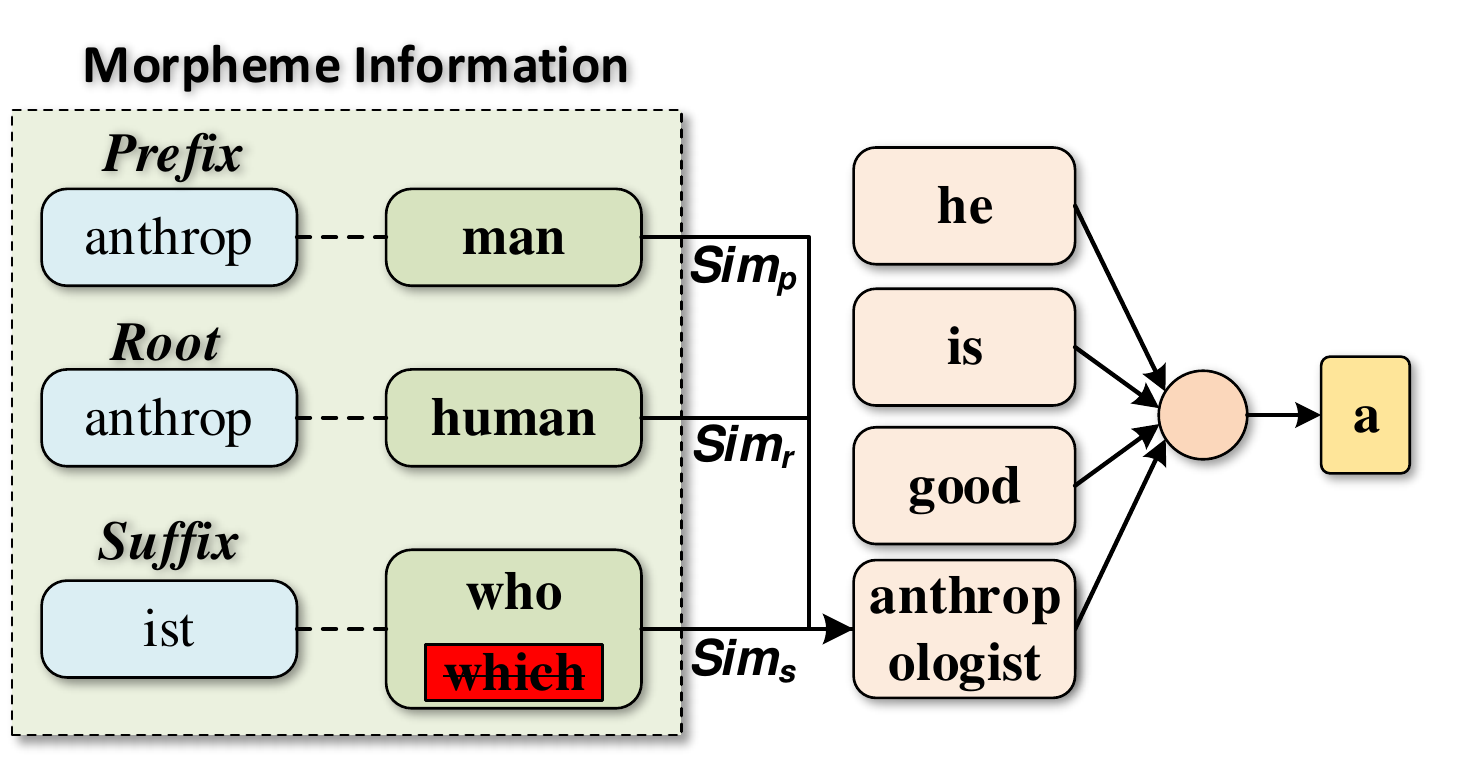}
	\caption{MWE-M. In this model, morpheme's meaning whose distance to the target word is the max will be added into the original representation of target word. For example, the suffix of ``anthropologist" has the meanings of ``who, which". According to the distance, only the meaning ``who" will be added to the representation of ``anthropologist".  }
\end{figure}

\subsection{Strategy to Match the Morpheme}
Till now, we need to discuss how to find the morphemes in a single token. Firstly, it is obvious that a number of words contain more than one prefix, root and suffix. For instance, ``anthropologist'' has the prefix ``a, an, anthrop''. The question raised here is which one should be used. Secondly, some morphemes' meanings of a word have nothing to do with that word. For instance, although ``apple" has the prefix ``a", it doesn't have the meanings of ``without, not". To solve these problems, some rules are defined as follows:
\begin{itemize}
\item When matching the morphemes of word $t_i$, the longest character sequence will be viewed as the morpheme.
\item We define a threshold $\lambda$. If the cosine distance between word $t_i$ and its morpheme $m_i$, ($m_i\in M_i$) is larger than $\lambda$, $m_i$ will be remained or abandoned otherwise. Based on cross validation, $\lambda$ is set as 0.4 in this paper.
\end{itemize}

\section{Experimental Setup}
Our models together with the baselines are tested on word similarity and syntactic analogy. 

\subsection{Corpus and Morphemes}
In this paper, we utilize a medium-sized English corpus to train all word embedding models. The corpus stems from the website of the 2013 ACL Workshop on Machine Translation\footnote{http://www.statmt.org/wmt13/translation-task.html} and is used in \cite{Kim2015Character}. This website lists a lot of corpora ordered by year when they are released. We choose the news corpus of 2009 whose size is about 1.7GB. It contains approximately 500 millions of tokens and 600 thousands of vocabularies. 
For better quality of all word embeddings, we filter all numbers and some punctuations out of the corpora. All morphemes used in this paper and their meanings are both collected from the website\footnote{https://msu.edu/~defores1/gre/roots/gre\_rts\_afx1.htm}. The morphemes set includes  90 prefixes, 241 roots and 64 suffixes. Based on our knowledge, the morphemes' meanings are refined. By using the lexicon PPDB \cite{ganitkevitch2013ppdb}, we create a map where the key is a word and the value is its morphemes' meanings set for convenience to scholars interested in this area.

\subsection{Baselines}
For comparison, we choose three state-of-the-art baselines including CBOW, Skip-gram \cite{mikolov2013efficient} and Glove \cite{pennington2014glove}. All baselines are explicit methods, and hold some state-of-the-art performance on the NLP tasks. 
For testing the differences between our models and traditional explicit ways, we directly utilize the morphemes rather than their meanings on the input layer of MWE-A. For clarity, this model is named explicitly morpheme-enhanced Word Embedding (EMWE) whose architecture is similar to the context-sensitive morphological recursive neural network (CSM-RNN) in \cite{luong2013better}. In this paper, we utilize the source code of word2vec\footnote{https://github.com/dav/word2vec} to train CBOW and Skip-gram. Glove is trained based on the code\footnote{http://nlp.stanford.edu/projects/glove/}. We modify the source of word2vec and train our models and EMWE.

\subsection{Parameter Settings}
Parameter settings have a great effect on the performance of word embeddings \cite{Levy2015Improving}. For fairness, all models are trained based on equal parameter settings. In order to accelerate the training process, CBOW, Skip-gram, EMWE together with our models are trained by using negative sampling. It is reported that the number of negative samples in the range 5-20 is useful for small corpus \cite{mikolov2013distributed}. If huge-sized corpus is used, the number of negative samples should be chosen from 2 to 5. According to the corpus we used, the number of negative samples is set as 20 in this paper. The dimension of word embedding is set as 200 like that in \cite{dhillon2015eigenwords}. We set the context  window size as 5 which is equal to the setting in \cite{mikolov2013distributed}.

\subsection{Evaluation Benchmarks}
We test all word embeddings from two aspects including word similarity and syntactic analogy.
\subsubsection{Word Similarity}
This experiment is used to evaluate word embedding's ability to capture semantic information from corpus. Each dataset is divided into three columns. The first two columns stand for word pairs and the last column is human score. On the task, there are two problems need to be solved. One is how to calculate the distance between two words. The other is how to evaluate the similarity between our results and human scores. For the first problem, we utilize the cosine distance to measure the distance between two words. This strategy is used in many previous works  \cite{mikolov2013distributed,pennington2014glove}. The second problem is solved via Spearman's rank correlation coefficient $(\rho)$. Higher $\rho$ means better performance. For English word similarity, we employ two golden standard datasets including Wordsim-353 \cite{finkelstein2001placing} and  RG-65 \cite{rubenstein1965contextual}. To avoid occasionality, however, we also utilize some other widely-used datasets including Rare-Word \cite{luong2013better}, SCWS \cite{Huang2012Improving},  Men-3k  \cite{bruni2014multimodal} and WS-353-Related \cite{agirre2009study}.  More details of these datasets are shown in Table 1.

\begin{table}[!th]
\small
\label{Wordsim}
\renewcommand\arraystretch{1.5}
\label{size}
\centering
\begin{tabular}{|c|c|c|c|}
\hline
Name & Pairs & Name & Pairs \\
\hline
RG-65 & 65 & RW & 2034\\
\hline
SCWS & 2003 & Men-3k & 3000\\
\hline
Wordsim-353 &353 & WS-353-REL & 252\\
\hline
\end{tabular}
\caption{Details of datasets. The columns of pairs indicate the number of word pairs in each dataset.}
\label{wordsimilarity}
\end{table}

\subsubsection{Syntactic Analogy}
Based on the learned word embedding, the core task of syntactic analogy is to answer the analogy questions ``$a$ is to $b$ as $c$ is to \_\_\_". In this paper, we utilize the Microsoft Research Syntactic Analogies dataset. This dataset with size of 8000 is created by Mikolov \cite{Mikolov2013Linguistic}. To answer the syntactic analogy questions ``$a$ is to $b$ as $c$ is to $d$" where $d$ is unknown, we assume that the word representations of $a$, $b$, $c$, $d$ are $v_a$, $v_b$, $v_c$, $v_d$, respectively. To get $d$, we firstly calculate $\widehat{v_d}=v_b-v_a+v_c$. Then, we find the word $\widehat{d}$ whose cosine distance to $\widehat{v_d}$ is the largest. Finally, we set $d$ as $\widehat{d}$. Formally, we describe it as following equation:
\begin{equation}
d=arg\max\limits_w\frac{v_w \widehat{v_d}}{\parallel v_w\parallel \parallel \widehat{v_d}\parallel}, w\in vocabulary
\end{equation}


\section{Results and Analysis}
\begin{table*}[!th]
\label{Wordsim-result}
\renewcommand\arraystretch{1.2}
\label{size}
\centering
\begin{tabular}{|c|c|c|c|c|c|c|c|}
\hline
&CBOW&Skip-gram&Glove&EMEW&MEW-A&MEW-S&MEW-M\\
\hline
Wordsim-353&58.77&61.94&49.40&60.01&62.05&\textbf{63.13}&61.54\\
RW&40.58&36.42&33.40&40.83&\textbf{43.12}&42.14&40.51\\
RG-65&56.50&62.81&59.92&60.85&62.51&62.49&\textbf{63.07}\\
SCWS&\textbf{63.13}&60.20&47.98&60.28&61.86&61.71&63.02\\
Men-3k&68.07&66.30&60.56&66.76&66.26&\textbf{68.36}&64.65\\
WS-353-REL&49.72&57.05&47.46&54.48&56.14&\textbf{58.47}&55.19\\
\hline
Syntactic Analogy&13.46&13.14&13.94&17.34&\textbf{20.38}&17.59&18.30\\
\hline
\end{tabular}
\caption{This table contains the results of word similarity and syntactic analogy. The numbers in bold indicate the best performance. All numbers are expressed as percentage (\%).}
\label{wordsimilarity}
\end{table*}

\subsection{Word Similarity}
Word similarity is conducted to test the semantic information which is contained in word embedding. From Table 2, we observe that our models surpass the comparative baselines on five datasets. It is remarkable that our models approximately outperform the base model CBOW for 5 percent and 7 percent on the golden standard Wordsim-353 and RG-65, respectively. On WS-353-REL, the difference between CBOW and MEW-S even reaches 8 percent. The advantage indicates the validity and effectiveness of our methods. We give some empirical explanations of the exciting promotion. Based on our strategy, more information will be captured in corpus. Like the example in Figure 2, the semantic information captured by CBOW only stems from sequence ``he is a good anthropologist". By contrast, in our models, not only the semantic information in the original sequence but also the information in sequence ``he is a good human" and ``he is a good man" are captured. Obviously, more semantic information means better performance. Based on the similar strategy, EMEW also performs better than other baselines but fail to get the performance as well as ours. Actually, EMEW alters the position of words in vector space to make the morpheme-similar words to group together, which interprets the better performance than CBOW. Nevertheless, like CBOW, EMEW can only capture the semantic information from original word sequence, which results in a little worse performance than ours. Specially, because of medium-size corpus and experimental settings, Glove doesn't perform as well as it in other papers \cite{pennington2014glove}.

\subsection{Syntactic Analogy}
In \cite{Mikolov2013Linguistic}, the dataset is divided into adjectives, nouns and verbs. In this paper, for brevity, we only report performance on the whole dataset. In Table 2, all of our models outperform the comparative baselines to a great extent. Compared with CBOW, the advantage of MEW-A even reaches to 7 percent. The result corresponds to our expectation. In Mikolov's dataset, we observe that the suffix of ``$b$" usually is the same as the suffix of  ``$d$" when answering question ``$a$ is to $b$ as $c$ is to $d$". Based on our strategy, morpheme-similar words will not only gather together but also have a trend to group near the morpheme's meanings, which makes our embeddings have the advantage to deal with the syntactic analogy problems. EMEW performs well on this task but is still weaker than our models as well. Actually, syntactic analogy is also a semantic-related task because ``$c$" and ``$d$" are with similar meanings. Based on the analysis of word similarity, our models are better to capture semantic information, which results in higher performance than EMEW's performance.


\subsection{Parameter Analysis}
It is obvious that parameter settings will affect the performance of word embedding. For example, the corpus with larger token size contains more semantic information, which can improve the performance on word similarity. In this paper, we analyze the effects of token size and window size on the performance of word embeddings.
\begin{figure}[h]
	\centering
	\includegraphics[width=1.0\linewidth]{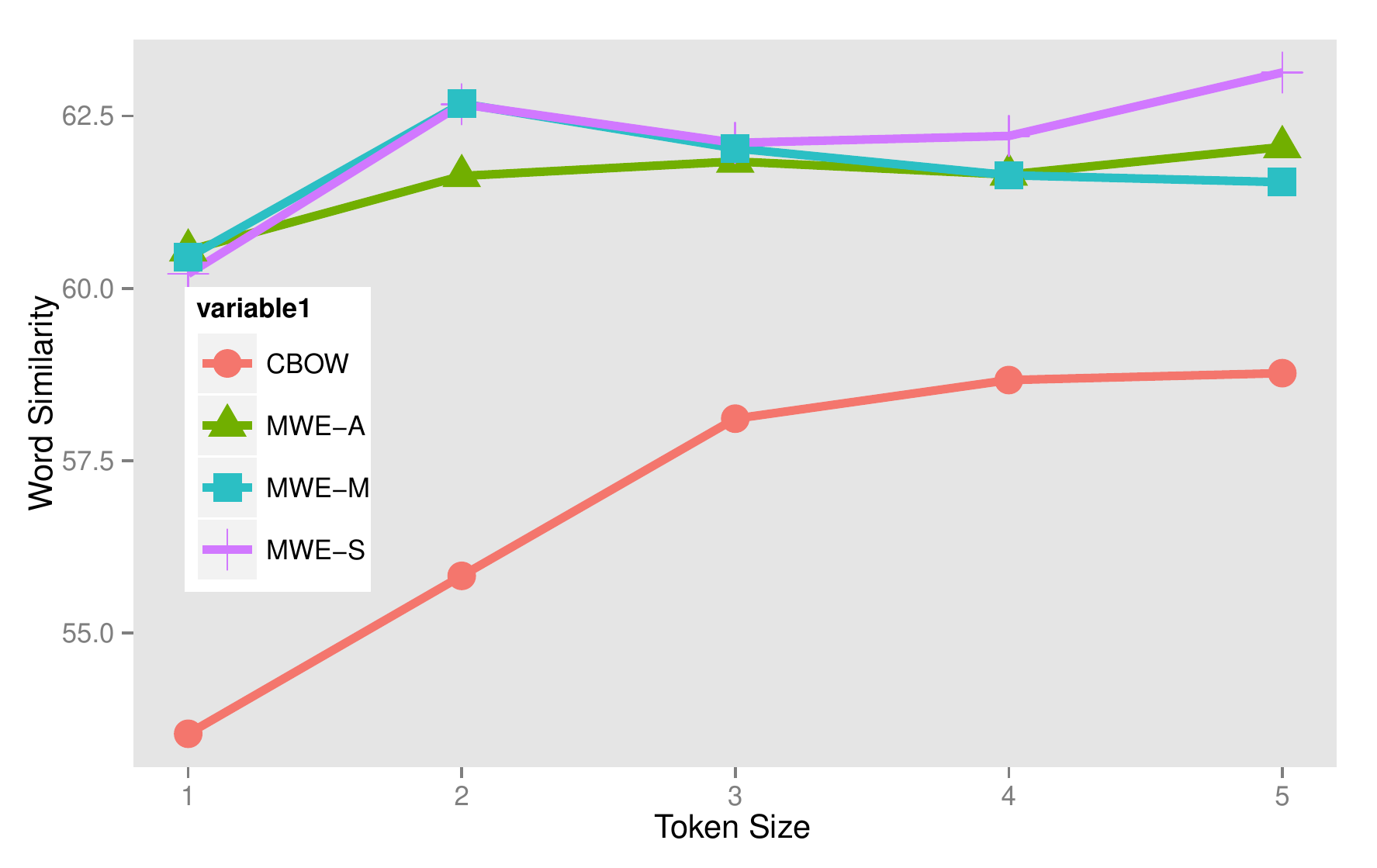}
	\caption{Parameter analysis of token size. X-axis and Y-axis indicate the token size and Spearman rank (\%) of word similarity respectively.}
\end{figure}

In the analysis of token size, we set the same parameter settings as what we did in the anterior section. The sizes of corpora used in this analysis are the 1/5, 2/5, 3/5, 4/5 and 5/5 of the corpus mentioned before, respectively. We utilize the result of word similarity on Wordsim-353 as the evaluation criterion. From Figure 5, we observe several phenomenons. Firstly,  the performance of our models is better than CBOW on each corresponding corpus, which indicates the effectiveness and validity of our methods. Secondly,  the performance of CBOW is sensitive to the token size. As a comparison, our models' performance seems more stable than CBOW. As we mentioned before,  in word similarity analysis, our methods can increase the semantic information of the corpus, which is a great property for dealing with some low-resource languages. It is worth noting that our performance on the smallest corpus is even better than CBOW's performance on the largest corpus.

In the analysis of window size, we observe that the performance of all word embeddings has a trend to ascend with the increasing of window size in Figure 6. Like the result in Figure 5, our models outperform CBOW under all pre-set conditions. The worst performance of our models is approximately equal to the best performance of CBOW, which may stem from the increased semantic information property of our methods.
\begin{figure}[h]
	\centering
	\includegraphics[width=1.0\linewidth]{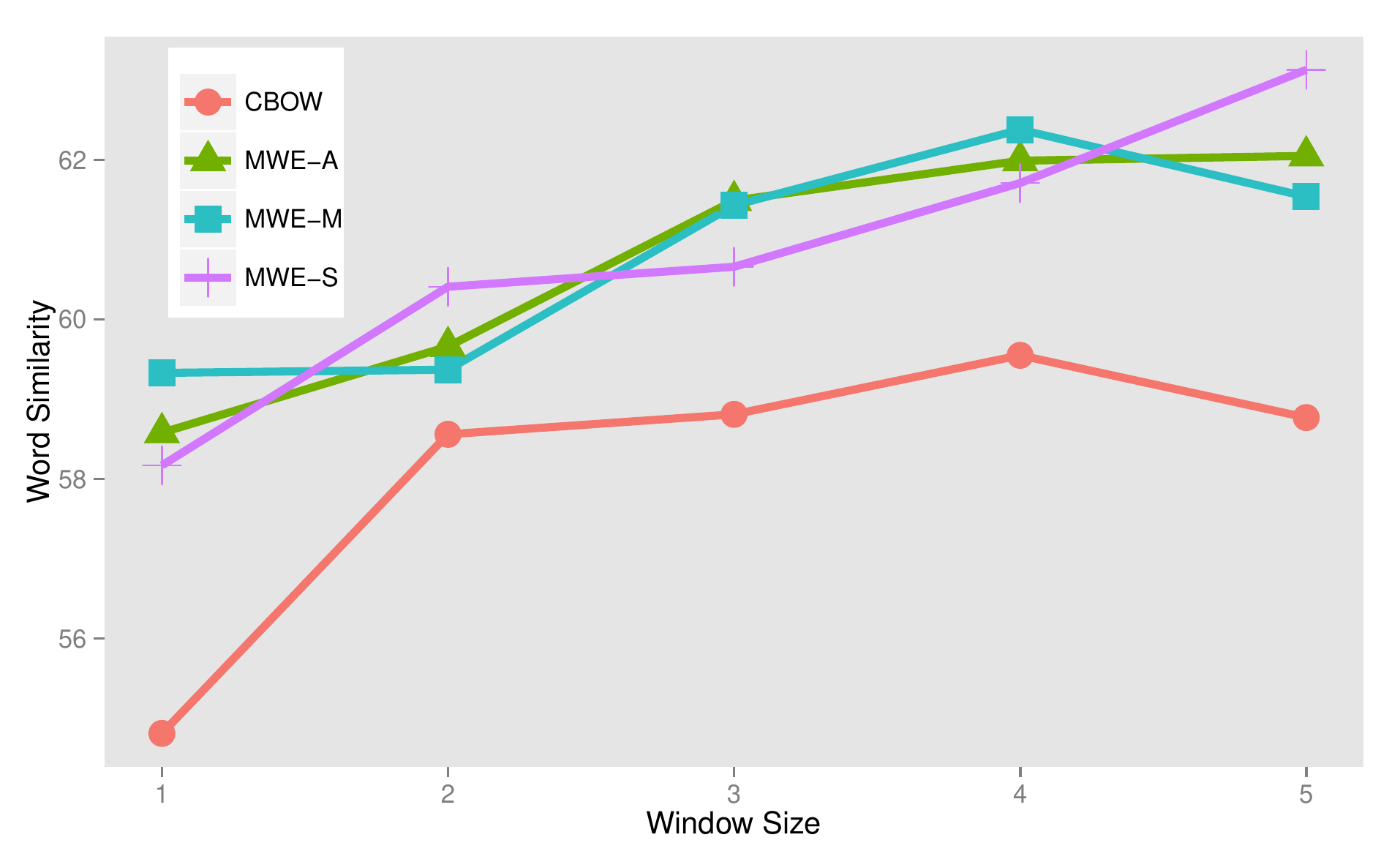}
	\caption{Parameter analysis of window size. X-axis and Y-axis indicate the window size and Spearman rank (\%) of word similarity respectively.}
\end{figure}

\subsection{N-nearest Words}
We randomly select 4 words which are with rich morpheme information from the vocabularies of CBOW and MWE-A respectively. Based on cosine distance, we show the top 3 nearest neighbors of these words in Table 3. CBOW's answers are always with similar meanings of target words but have different morphemes. However, MWE-A always gives the results which not only are near to the target words but also have similar morphemes. Moreover, the results of MWE-A seems better than CBOW's answers.
\newcommand{\tabincell}[2]{\begin{tabular}{@{}#1@{}}#2\end{tabular}}
\begin{table}[h]
\scriptsize
\centering
\begin{tabular}{|p{0.22\columnwidth}|p{0.25\columnwidth}|p{0.25\columnwidth}|}
  \hline
    & \hspace*{\stretch{1}}\textbf{CBOW}\hspace*{\stretch{1}} & \hspace*{\stretch{1}}\textbf{MWE-A}\hspace*{\stretch{1}} \\
  \hline
  international & \tabincell{l}{hemispheric, global,\\ nongovernmental}  &  \tabincell{l}{regional, global,\\ nongovernmental} \\
  \hline
  postretirement & \tabincell{l}{OPEB, noncurrent,\\ accruals} & \tabincell{l}{postemployment, \\OPEB, noncurrent} \\
  \hline
  pharmacodynamic & \tabincell{l}{pegylated, protease,\\ brostallicin} & \tabincell{l}{pharmacokinetic, \\pharmacokinetics,\\ elagolix} \\
  \hline
  inequitable & \tabincell{l}{mispricing,\\ penalising,\\ constraining} & \tabincell{l}{irresponsible,\\ unsustainable,\\ unfair} \\
  \hline
\end{tabular}
\caption{The results of N-nearest words. We give the top 3 nearest words of the target words.}
\end{table}

To further demonstrate the validity and effectiveness of our models, the dimension of word embedding is reduced from 200 to 2 with Principal Component Analysis (PCA). We randomly select several words from the embedding of MWE-A and illustrate them in Figure 7. Different colors stand for words with different morphemes. It is apparent that words with similar morpheme have a trend to group together.
\begin{figure}[h]
	\centering
	\includegraphics[width=1.0\linewidth]{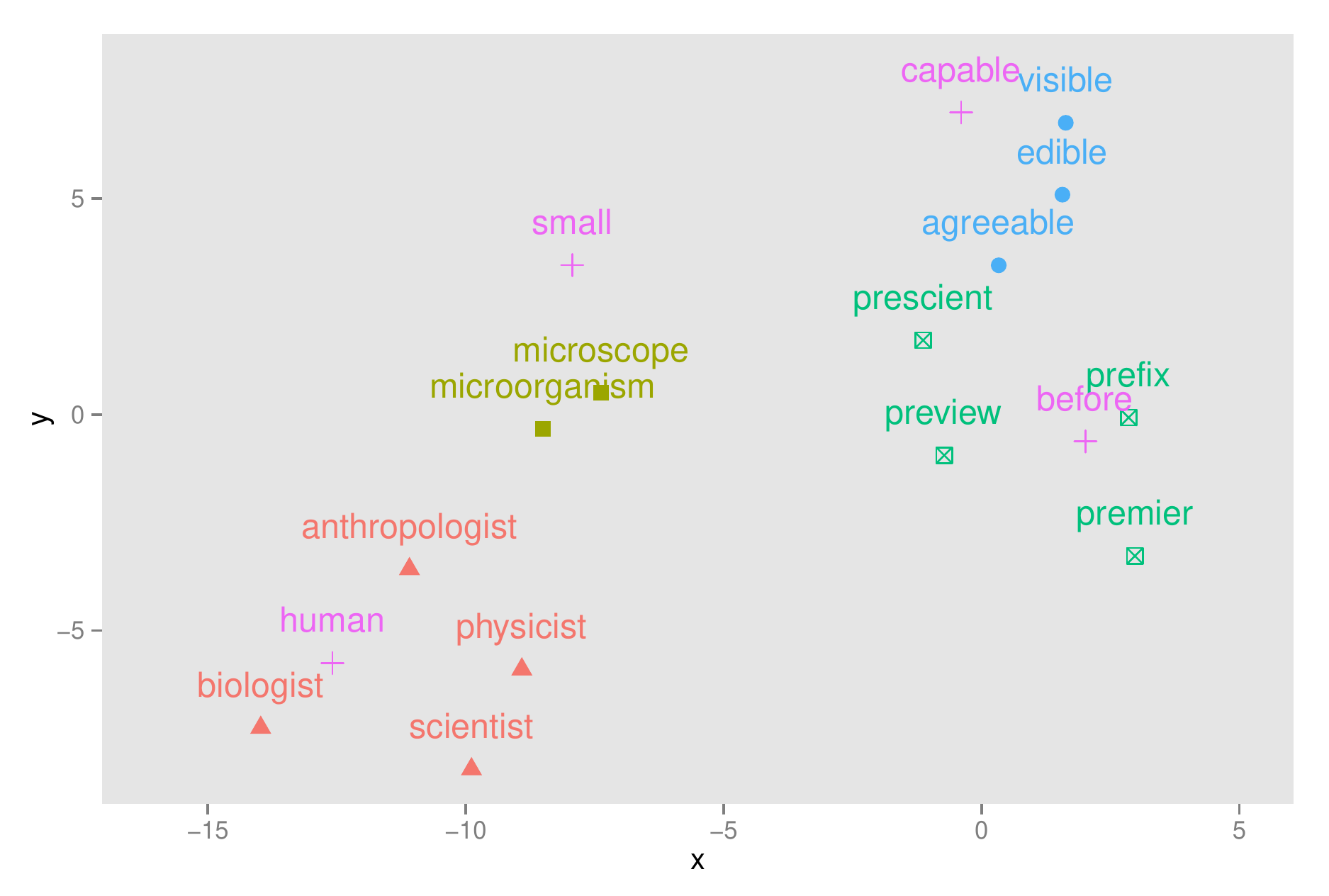}
	\caption{Validity of our methods. Based on PCA, we randomly select several words from word embedding of MWE-A and illustrate them in this figure. ``+" indicates the morphemes' meanings and morpheme-similar words tend to group together.}
\end{figure}

\section{Conclusion}
In this paper, we proposed a novel method to implicitly incorporate morphological information into word embedding. In our models, the morphemes' meanings are added on the input layer of CBOW rather than morphemes themselves. Based on this strategy, morpheme-similar words will not only gather together but also have a trend to group near their morphemes' meanings. Based on the different strategies to add the morphemes' meanings, three models named MWE-A, MWE-S and MWE-M were built. To test the performance of our embeddings, we utilized three comparative baselines and tested them on the word similarity and syntactic analogy tasks. The results show that our models outperform the baselines on five word similarity datasets. On the golden standard Wordsim-353 and RG-65, our models even approximately outperform CBOW for 5\% and 7\%, respectively. On syntactic analogy task, our models surpass the baselines to a great extent. Compared with CBOW, MWE-A approximately outperforms it for 7\%. Compared to the explicit method, our models also show a great advantage.

Significant	improvements on both tasks lead us to believe the validity of our models. Last but not the least, our models seem to have great advantages to train some low-resource but morpheme-rich languages according to the observation of token size analysis. In future work, we will test our models' ability to deal with some morpheme-rich languages like Germany and French.


\end{document}